# Chemical-protein relation extraction with ensembles of SVM, CNN, and RNN models

Yifan Peng[1], Anthony Rios[1,2], Ramakanth Kavuluru[2,3], Zhiyong Lu[1]

[1]National Center for Biotechnology Information, National Library of Medicine, National Institutes of Health, Bethesda, MD,
[2]Department of Computer Science, University of Kentucky, Lexington, KY
[3]Division of Biomedical Informatics, Department of Internal Medicine, University of Kentucky, Lexington, KY

*Abstract*—Text mining the relations between chemicals and proteins is an increasingly important task. The CHEMPROT track at BioCreative VI aims to promote the development and evaluation of systems that can automatically detect the chemical-protein relations in running text (PubMed abstracts). This manuscript describes our submission, which is an ensemble of three systems, including a Support Vector Machine, a Convolutional Neural Network, and a Recurrent Neural Network. Their output is combined using a decision based on majority voting or stacking. Our CHEMPROT system obtained 0.7266 in precision and 0.5735 in recall for an f-score of 0.6410, demonstrating the effectiveness of machine learning-based approaches for automatic relation extraction from biomedical literature. Our submission achieved the highest performance in the task during the 2017 challenge.

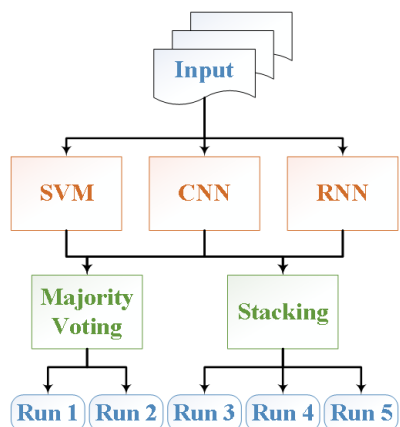

Fig. 1 Architecture of the systems for the CHEMPROT task

## I. INTRODUCTION

Recognizing the relations between chemicals and proteins is crucial in various tasks such as precision medicine, drug discovery, and basic biomedical research. Biomedical researchers have studied various associations between chemicals and proteins and published their findings in biomedical literature. While manually extracting chemical-protein relations from biomedical literature is possible, it is often costly and time-consuming. Alternatively, text mining methods could automatically detect these relations effectively. The BioCreative VI track 5 CHEMPROT task[1] aims to promote the development and evaluation of systems that are able to automatically detect in running text (PubMed abstracts) relations between chemical compounds/drug and genes/proteins. In this paper, we describe our approaches and results for this task.

## II. METHODS

In the CHEMPROT track, the organizers developed a chemical-protein relation corpus composed of 4,966 PubMed abstracts, which were divided into a training set (1,020 abstracts), development set (612 abstracts) and test set (8,00 abstracts).

Unlike other relation corpora (1, 2), cross-sentence relations are rare in the corpus, appearing only in less than 1% in the training set. We also noticed that some chemical-protein pairs have multiple labels, but they only appear 10 times in the training set. As a result, our system treated the relation extraction task as a multiclass classification problem, and to simplify the problem, our system only focuses on the chemical-protein relations occurring in a single sentence.

We addressed the CHEMPROT task using two ensemble systems that combine the results from 3 individual models, similar to our previous BioCreative submissions (3). An overview of the system architecture is shown in Figure 1. The individual systems included are a Support Vector Machine (SVM), a Convolutional Neural Network (CNN), and a Recurrent Neural Network (RNN) (4-6). We will describe these models together with the ensemble algorithms in the following subsections.

### A. Rich Feature SVM

In our SVM system, the following features are exploited.

Words surrounding the chemical and gene mentions of interest: These features include the lemma form of a word, its part-of-speech tag, and chunk types. We used the Genia Tagger to extract the features (7). The window size is 5.

Bag-of-words between the chemical and gene mentions of interest in a sentence: These features include the lemma form of a word and its relative position to the target pair of entities (before, middle, after).

---
[1] http://www.biocreative.org/tasks/biocreative-vi/track-5/

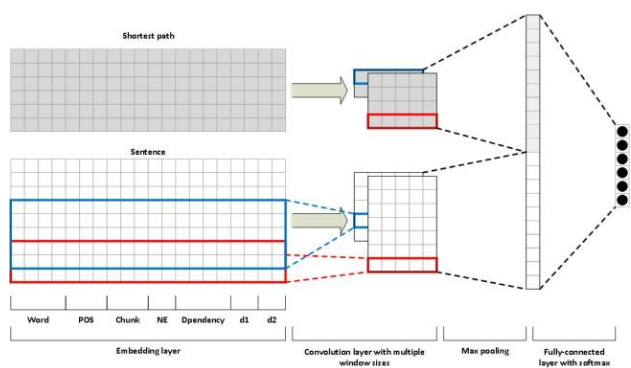

Fig. 2 Overview of the CNN model

The distance (the number of words) between two entity mentions in a sentence.

The existence of a keyword between two mentions often implies a specific type of a relation, such as "inhibit" for relation "CPR:4" and "agonism" for relation "CPR:5". Therefore, we manually built the keyword list from the training set and used them as features as well.

Shortest-path features include vertex walks (v-walks) and edge walks (e-walks) on the target pair in a dependency parse graph (8). An e-walk includes one word and its two dependencies. A v-walk includes two words and their dependency. For example, the shortest path of <Gemfibrozil, nitric-oxide synthase> extracted from the sentence "Gemfibrozil$_{CHEMICAL}$, a lipid-lowering drug, inhibits the induction of nitric-oxide synthase$_{GENE-N}$ in human astrocytes." is "Gemfibrozil ← nsubj ← inhibits → dobj → induction → nmod:of → nitric-oxide synthase". Thus the e-walks are "nsubj – inhibits – dobj" and "dobj – induction – nmod:of – nitric-oxide synthase". The v-walks are "Gemfibrozil – nsubj – inhibits", "inhibits – dobj – induction", and "induction – nmod:of – nitrix-oxide synthase".

We trained the SVM system using a linear kernel[2]. In our submissions, the penalty parameter C is 1 and the tolerance for stopping criteria is 1e-3. We also balanced the feature instances by adjusting their weights inversely proportional to class frequencies in the training set. We used one-vs-rest multiclass strategy.

*B. Convolutional Neural Networks*

We followed the work of Peng and Lu (9) to build the CNN model. Instead of using multichannels, we applied one channel but used two input layers (Fig. 2). One is the sentence sequence and the other is the shortest path between the pair of entity mentions in the target relation.

In our model, each word in either a sentence or a shortest path is represented by concatenating embeddings of its words, part-of-speech tags, chunks, named entities, dependencies, and positions relatively to two mentions of interest. The pre-trained word embedding vectors was learned on PubMed articles using the gensim word2vec implementation with the dimensionality set to 300 (10). The part-of-speech tags, chunks, and named entities were obtained from Genia Tagger (7).

[2] http://scikit-learn.org/

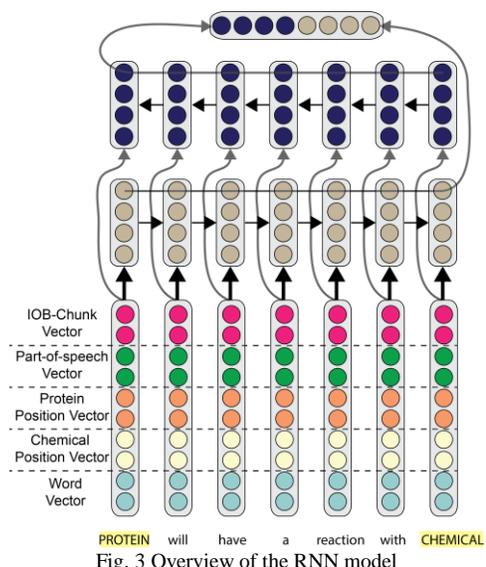

Fig. 3 Overview of the RNN model

The dependency information was obtained using the Bllip parser with the biomedical model and the Stanford dependencies converter (11-13). For each word, we used the dependency label of the "incoming" edge of that word in the dependency graph.

For each input layer, we applied convolution to inputs to get local features with window sizes of 3 and 5. ReLU function used as the activation unit and 1-max pooling was then performed to get the most useful global feature from the entire sentence.

In the fully connected layer, we concatenate the global features from both the sentence and the shortest path and then applied a fully connected layer to the feature vectors and a final softmax to classify the 6 classes (5 positive + 1 negative). We also used the dropout technique to prevent overfitting.

All parameters were trained using the Adam algorithm to optimize the cross entropy loss on a mini-batch with a batch size of 32 (14).

*C. Recurrent Neural Netwoks*

For our RNN model, we build on the work of Kavuluru et al. (15). Specifically, we train a bi-directional long-short term-memory (Bi-LSTM) recurrent model (Fig. 3), where the input to the model is a sentence. In this work, we don't consider the character level Bi-LSTM.

Similar to our CNN model, we concatenate the word embedding with the part-of-speech, IOB-chunk tag, and two position embeddings. The two position embeddings represent the relative location of the word with respect to the two entity mentions. It is important to note that we update the embeddings (word, part-of-speech, chunk, and position) during training.

After passing a sentence through our Bi-LSTM model, we obtain two hidden representations for each word. One representing the forward context, and the other representing the backward. We concatenate the two representations to obtain the final representation of each word. To obtain a representation of the sentence, we use max-over-time (1-max) pooling across hidden state word representations.

Next, we pass the max-pooled sentence representation to a fully connected output layer. Unlike our CNN, we only apply a linear transformation without a softmax operation. Furthermore, the output layer only has 5 classes, where we completely discard the negative class. Specifically, we use the pairwise ranking loss proposed by Santos et al (16). Intuitively, the negative class will be noisy compared to the 5 positive classes. Rather than learning to predict the negative class explicitly, we force the 5 outputs to be negative. At prediction time, if all 5 class scores are negative, then we predict the negative class. Otherwise, we predict the class with the largest positive score.

Before training our model, we preprocess the dataset by replacing each word in the corpus that occurs less than 5 times with an unknown (UNK) token. Also, given each instance is comprised of a sentence and two entity mentions, we replace each entity with the tokens CHEMICAL or PROTEIN dependent on what the specific mention represents.

Finally, we train our RNN model using the Adam optimizer with a mini-batch size of 32. For the Adam optimizer, we set the learning rate to 0.001, beta1 to 0.9 and beta2 to 0.999. In addition, we apply recurrent dropout of 0.2 in the Bi-LSTM model and standard dropout of 0.2 between the max-pooling and output layers. We use pre-trained word vectors (6B Token GLOVE[3]) with a dimensionality of 300. Likewise, the POS, position, and chunk vectors are randomly initialized, and each has a dimensionality of 32. We should note that both the POS and chunk tags are extracted using NLTK[4]. Lastly, we set the hidden state size of the LSTM models to 2048.

### D. A Majority Voting System

In the voting system, we combined the results of the three models using a majority voting. That is, we select the relations that are predicted by more than 2 models.

### E. A Stacking System

While voting is a straightforward way to combine our SVM, CNN, and RNN models, methods that are more sophisticated can improve our performance. Specifically, we use stacking to combine the predictions of each model. Stacking works by training multiple base models (SVM, CNN, and RNN), then trains a meta-model using the base model predictions as features.

For our meta-model, we train a Random Forest (RF)[5] classifier. First, in order to train the RF, we capture the scores for each class from all 3 models on the development set. In total, we have the following 17 features: 6 from the SVM, 6 from CNN, and 5 from the RNN (because we used a ranking loss).

For the CNN scores, we use the unnormalized scores for each class before passing them through the softmax function. Finally, we train the RF on the development set using 50,000 trees and the gini splitting criteria.

## III. RESULTS AND DISCUSSION

In the CHEMPROT track, the test set contains 800 abstracts.

Our submissions were prepared with an ensemble of 3 models. We built every SVM, CNN and RNN model using 80% total data (training + development) and built the ensemble system using the remaining 20% of the total data. To reduce variability, 5-fold cross-validation was performed using different partitions of the data. As a result, we obtained 5 SVMs, 5 CNNs, and 5 RNNs in total. We submitted 5 runs as our final submissions. Runs 1 and 2 use a majority voting system and Runs 3-5 use a stacking system. Each run uses one SVM, CNN, and RNN from one cross validation iteration.

Table 1 shows the overall performance of the ensemble system as reported by the organizer, where 'P', 'R', 'F' denotes precision, recall, and F1 score, respectively.

TABLE I. RESULTS FOR OUR ENSEMBLE SYSTEM ON TEST SET

| Run | System | P | R | F |
|---|---|---|---|---|
| 1 | Majority Voting | 0.7311 | 0.5685 | 0.6397 |
| 2 | Majority Voting | 0.7266 | 0.5735 | 0.6410 |
| 3 | Stacking | 0.7437 | 0.5529 | 0.6343 |
| 4 | Stacking | 0.7283 | 0.5503 | 0.6269 |
| 5 | Stacking | 0.7426 | 0.5382 | 0.6241 |

## IV. CONCLUSION

In this manuscript, we describe our submission in the BioCreative VI CHEMPROT task. The results demonstrate that our ensemble system can effectively detect the chemical-protein relations from biomedical literature. We also show that the domain specific features are useful for in this task.

### ACKNOWLEDGMENT

The authors thank the organizers of the BioCreative VI CHEMPROT task. This research is supported by the Intramural Research Programs of the National Institutes of Health, National Library of Medicine. A.R. was a summer intern at the NCBI/NIH and supported by the NIH Intramural Research Training Award. R.K.'s involvment is supported by the National Library of Medicine through grant R21LM012274.

---

[3] https://nlp.stanford.edu/projects/glove/
[4] http://www.nltk.org/
[5] http://scikit-learn.org